# RNN-based Online Handwritten Character Recognition Using Accelerometer and Gyroscope Data

**Davit Soselia, Shota Amashukeli, Irakli Koberidze, Levan Shugliashvili**

San Diego State University, San Diego, CA USA,

{dsoselia, samashukeli, ikoberidze, ldadiani58}@sdsu.edu

## Abstract

This abstract explores an RNN-based approach to online handwritten recognition problem. Our method uses data from an accelerometer and a gyroscope mounted on a handheld pen-like device to train and run a character prediction model. We have built a dataset of timestamped gyroscope and accelerometer data gathered during the manual process of handwriting Latin characters, labeled with the character being written; in total, the dataset consists of 1500 gyroscope and accelerometer data sequences for 8 characters of the Latin alphabet from 6 different people, and 20 characters, each 1500 samples from Georgian alphabet from 5 different people. with each sequence containing the gyroscope and accelerometer data captured during the writing of a particular character sampled once every 10ms. We train an RNN-based neural network architecture on this dataset to predict the character being written. The model is optimized with categorical cross-entropy loss and RMSprop optimizer and achieves high accuracy on test data.

## Introduction

Recent developments in deep learning have led to rapid progress in the field of handwritten digit and character recognition. In particular, standardized and ubiquitous image-based handwritten character datasets such as EMNIST and deep convolutional neural network architectures have led to new state-of-the-art performances in the task of offline handwritten character recognition. In contrast, the task of online handwritten character recognition using non-image data gathered from a pointing device is still largely performed through traditional statistical learning methods and shallow neural network architectures.

While various pen-based devices have been proposed (Zhang, Yuan, and Zhang 2008), they require either special surfaces, or broad strokes (Tunçer, 2016), or are mainly focused on Arabic numeral recognition (Wang, Hsu, and Chu 2013). This project proposes incorporating RNNs to achieve higher accuracy without dependence on special writing surfaces or movements.

## Device

The device consists of a microcontroller (STM32-F1), 3 axes accelerometer/gyroscope (MPU-6050), Bluetooth transmitter (CSR-BC417) and a push button, the setup is similar to the one introduced in Online handwriting recognition using an accelerometer-based pen device (Wang, Hsu, and Chu 2013). When the user starts writing the push button is pressed and microcontroller starts retrieving data from the accelerometer/gyroscope with 10ms intervals. Once the button is released, the data is sent from microcontroller to Bluetooth transmitter and then to the computer which is paired with the device via Bluetooth, where the framework makes predictions.

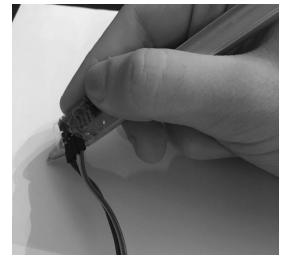

*Image of the prototype attached to a regular pen*

## Dataset and Preprocessing

The accelerometer provides acceleration magnitude in X Y and Z axis every 10ms. and gyroscope provides angular Velocity in 3D space. The data is stored in a 6xT matrix where T is the number of timesteps. The positioning of the index finger was controlled to acquire similar orientation. 20:80 division was made for testing and training.

The augmentation is done using sample averaging and random noise introduction. Random noise is introduced using Gaussian distribution. The data is then scaled to the [-1,1] range. The averaging passes a window of 6xN over the input matrix in M steps, where N is the size of the window and M is the stride.

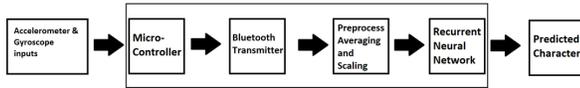

*The full process of the character recognition*

## Model

RNN model with 2 LSTM and 2 dense layers is used for the character classification. The input shape is 1xNx6 where N is the number of timesteps. The LSTM layers have 20, and 25 units respectively. Hard Sigmoid function is used for activation. The dense layers have 25 units and use ReLU activation.

## Training and Testing

Training and testing are conducted in three ways:
- train on the handwriting of a group of 4 people, test with 2 different people.
- train and test on the handwriting of the same group of 6 people.
- train on the handwriting 4 people, test on 4 plus 2 unknown handwritings.

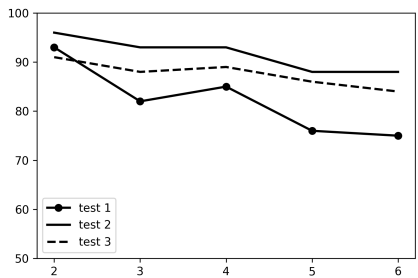

*Figure 1. Correlation of accuracy on the number of classes for a fixed number of training samples for each class.*

As seen in Figure 1 the accuracy for the whole testing set in each case is as follows 75, 88, 84 While the binary classification shows the highest accuracy of 91, 96, 93 for each case. As Figure 2 shows, the drop of accuracy can be compensated with the increase in the train set size. The model is tested for real-world use by passing recognized output through autocorrection software. Latest autocorrect methodologies (Ghosh and Kristensson 2017) increase overall word accuracy to 91%.

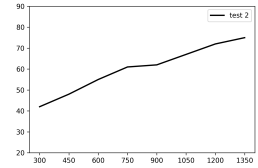

*Figure 2*

## Conclusion

The abstract proposes a system for recognizing handwritten characters using acetometer based pen device with high (over 88%) accuracy. We build a dataset of Latin and Georgian characters. The dataset consists of 1500 gyroscope and accelerometer data sequences for select characters of the Latin and Georgian alphabet. RNN is successfully employed to find unique character patterns among various handwriting styles of multiple individuals. This technology could greatly increase the efficiency and decrease the cost of digital backups

Further steps include creating a large-scale database with various handwritten samples, for different handwriting styles, and left and right-handed users, as well as further testing the system in the real-world applications.

## Acknowledgements

We express our gratitude to Dr. Magda Tsintsadze and SDSU-Georgia program for supporting this research

## References

Ghosh, S., and Kristensson P. O. 2017. Neural Networks for Text Correction and Completion in Keyboard Decoding. In *ArxXiv preprint,* arxiv:1709.06429.

Tunçer, E. 2016. Accelerometer Based Handwritten Character Recognition Using Dynamic Time Warping, Master's thesis, İzmir Institute of Technology.

Wang, J. S., Hsu, Y. L. and Chu, C. L., 2013. Online handwriting recognition using an accelerometer-based pen device. In *2nd International Conference on Advances in Computer Science and Engineering* pp.229-232

Zhang, S., Yuan, C. and Zhang, Y., 2008, July. Handwritten character recognition using orientation quantization based on 3D accelerometer. In *Proceedings of the 5th Annual International Conference on Mobile and Ubiquitous Systems: Computing, Networking, and Services* (p. 54). ICST (Institute for Computer Sciences, Social-Informatics and Telecommunications Engineering).